# Bio-inspired data mining: Treating malware signatures as biosequences

## Ajit Narayanan[*] and Yi Chen

School of Computing & Mathematical Sciences, Auckland University of Technology, Auckland, New Zealand.

**ABSTRACT**

The application of machine learning to bioinformatics problems is well established. Less well understood is the application of bioinformatics techniques to machine learning and, in particular, the representation of non-biological data as biosequences. The aim of this paper is to explore the effects of giving amino acid representation to problematic machine learning data and to evaluate the benefits of supplementing traditional machine learning with bioinformatics tools and techniques. The signatures of 60 computer viruses and 60 computer worms were converted into amino acid representations and first multiply aligned separately to identify conserved regions across different families within each class (virus and worm). This was followed by a second alignment of all 120 aligned signatures together so that non-conserved regions were identified prior to input to a number of machine learning techniques. Differences in length between virus and worm signatures after the first alignment were resolved by the second alignment. Our first set of experiments indicates that representing computer malware signatures as amino acid sequences followed by alignment leads to greater classification and prediction accuracy. Our second set of experiments indicates that checking the results of data mining from artificial virus and worm data against known proteins can lead to generalizations being made from the domain of naturally occurring proteins to malware signatures. However, further work is needed to determine the advantages and disadvantages of different representations and sequence alignment methods for handling problematic machine learning data.

All data, machine learning and biological tools used in this paper are publicly available and free. Computer malware signatures were downloaded from VX heavens: www.vxheavens.com. The multiple sequence alignment techniques were ClustalW and T-Coffee from EBI: www.ebi.ac.uk/Tools/msa/tcoffee. Various data mining functions within Weka (Waikato Environment for Knowledge Analysis) were used for machine learning involving cross-validation, rule extraction and classification: www.cs.waikato.ac.nz/ml/weka. Biological match of consensuses and meta-signatures was checked through the PRINTS database available via Motif3D at the University of Manchester http://www.bioinf.manchester.ac.uk/dbbrowser/motif3d/motif3d.html and FingerPRINTScan at the EBI (http address above). QUARK and the Protein Data Bank were used for consensus modeling and protein matching: zhanglab.ccmb.med.umich.edu/QUARK/ and www.pdb.org/pdb/home/home.do, respectively. The National Centre for Biotechnology Information (NCBI) was used for checking the biological function of matched proteins: ncbi.nlm.nih.gov/. The commercial package SPSS v19 was used for statistical analysis of alignment lengths and accuracy results by T tests and analysis of variance.

[*] Corresponding author: Ajit Narayanan, Ajit.Narayanan@aut.ac.nz. Tel: +64 9921 9345; Fax: +64 9921 9944

# 1 INTRODUCTION

Such is the growth of malware (the generic term given to any program or code intended to cause disruption or gain access to unauthorised information and resources) that Symantec reported over 3 billion malware attacks in 2010, with 93% increase in web attacks (Symantic Internet Security Threat Report, 2011). The traditional method for dealing with viruses and worms (two of the most common types of malware) is to use anti-viral software (AVS) that looks for 'signatures' which represent critical parts of executable code in numerical form. Whereas a virus requires some action to be propagated, such as being attached to a program by the user, a worm can propagate by itself. The aim of virus and worm writers is usually to cause some damage to computer systems, hence the term 'malware'.

Signatures can be calculated from a pattern of operations in the malware code or can represent the encryption algorithm used to hide the virus or worm. Signatures were originally and continue to be identified and calculated by human experts, and are typically a sequence of hexadecimal numbers intended to uniquely identify viruses and worms. Automatic generation of signatures for new malware continues to be a difficult problem (Tang and Chen, 2008). Such signatures can also be consistent for a 'family' of viruses or worms that share parts of the code or have similar function and are essentially variants of each other. For instance, 'Virus.Acad.Bursted.a' is a typical computer virus name that indicates the platform (Autocad, or 'Acad'), the family (Bursted) and the variant 'a'. Achieving consistency of signatures for members of the same family is especially important when dealing with polymorphic (the functional parts of the code are the same but hidden differently) and metamorphic (the function remains the same but the code is altered with every replication) malware designed to avoid such signature detection (Szor, 2005; Parikka, 2007). Due to the security dangers inherent in making the original malware code available for public dissemination, only signatures are made publicly available.

AVS scanners use a dictionary or library of signatures in a variety of different ways. For instance, for simple polymorphic malware detection the hexadecimal representation of a signature can be used to match against incoming network packets containing bytes also represented in hexadecimal. This allows the AVS to check for contiguous similarities between parts of the signature and packet contents. For metamorphic and more complex polymorphic malware detection, increasingly sophisticated techniques must be used that allow for contiguous parts of the signature to be detected non-contiguously across different packets (Bayoglu and Sogukpinar, 2008). Signature detection through pattern matching is usually supported by other techniques, such as stateful monitoring, to minimize false positives and false negatives (Chen, 2004).

Malware writers adopt a variety of sophisticated techniques for avoiding detection. By the time a new variant is identified and signatures released, the infection may already have reached epidemic proportions (Strickland, 2011). One of the problems in applying automatic data mining techniques to malware code directly, even if it is available, is the variable length of the code (Xinguang *et al*., 2009), since most data mining and other machine learning techniques assume fixed length sequences with a column representing measurements of the same variable across many samples. There is surprisingly little work reporting on the application of machine learning techniques to malware signature detection, mainly due to the problem of obtaining malware source code as well as the need to deal with variable length code to identify the critical parts of the code from which to derive signatures. The little work that exists in machine learning focuses on anomalous behavior detection (e.g. Rieck *et al*., 2008; Singhal and Raul, 2012) rather than code or signature analysis. Also, mining the signatures directly can lead to results that are difficult to interpret, since the hexadecimal signatures cannot always be mapped back to meaningful and individual operations in the source code (op code). The variable length of the malware code, the difficulty of legally obtaining source malware code for detailed analysis, the lack of interpretability of results if hexadecimal signatures are used and the partially sequential aspects of the data all obstruct the use of machine learning techniques, thereby limiting their use in the urgent problem of finding automatic ways of generating signatures.

In bioinformatics, sequence analysis is used to understand the relationship between two (pairwise alignment) or more (multiple sequence alignment) sequences of genetic information, such as DNA or amino acids (Mount, 2001). In particular, sequence analysis and alignment can be used to identify conserved regions or motifs (regions of similarity) in biological data that identify common genes and shared ancestry as well as shed light on structure and function (Needleman and Wunsch, 1970; Smith and Waterman, 1981). Sequence alignment techniques are not confined to biological sequences and there have been applications of sequence alignment in linguistics (Kondrak, 2002) and marketing (Prinzie and Van den Poel, 2006). The first demonstration of the use of alignment for detecting computer viruses was reported in 2007 (McGhee, 2007) and used profile hidden Markov models (Eddy, 1998) constructed from position specific information for scoring. However, only a subset of possible op codes was used for alignment, resulting in the exclusion of many viral op codes deemed not important. Also, the HMM worked not on a biological representation of the op codes but on an alphabet that was neither DNA nor amino acid.

Earlier work (Chen *et al*., 2012a; Chen *et al*., 2012b) reported that traditional symbolic data mining (RIPPER (Cohen, 1995) and C4.5 (Quinlan, 1986)) in Weka (known as JRip and J48, respectively) had returned 60% accuracy using 10-fold cross-validation on 60 signatures (30 virus, 30 worm) without alignment. This confirmed the problematic nature of signature analysis. After conversion to amino acid representation followed by standard multiple alignment, classification accuracy increased significantly, allowing hexadecimal meta-signatures to be extracted for distinguishing virus from worm signatures. But only one amino acid representation of signatures was tried and only one alignment method chosen. Also, only a couple of machine learning techniques were tried. This leads to questions about the reliability of earlier results and the applicability of the methods for larger sample sets. One of the aims of this paper is to double the number of signatures and to explore the implications of adopting different predefined substitution matrices alongside two different alignment techniques with three different residue representations. Another aim is to increase the variety of machine learning tools for evaluating the effects of different representation on classification and prediction from two to five.

It was not possible to attach meaning to the previously extracted meta-signatures (Chen *et al*., 2012a; Chen *et al*., 2012b), since the original op code that gave rise to the signatures was not available. As we shall see, the preserved or conserved results of alignment (i.e. the consensus sequence of letters that are similar across sequences) can be given biological realization through protein modeling to help in the evaluation of different representations as well as provide some interpretability of the results. Such biological realizations could provide new, biologically inspired ways of looking at malware signatures for future AV dictionaries to use.

While the application of machine learning techniques (e.g. neural networks, genetic algorithms, rule extractors) in bioinformatics is well established (e,g, Baldi and Brunak, 2001; Keedwell and Narayanan, 2005), there is less understanding of how bioinformatics techniques can be applied to machine learning, especially when the data being mined is problematic from a machine learning perspective. The paper contains two sets of systems and methods and therefore two sets of results. Standard alignment methods and machine learning techniques are applied to data represented as biosequences in the first set of systems, methods and results. Standard bioinformatics protein database searching and protein modeling techniques are applied in the second. The combined results will lead to a discussion of the advantages and disadvantages of treating machine learning data as amino acid sequences.

To summarize, the research issues being addressed in this paper are: (a) the reliability of representing problematic data as biosequences followed by alignment and machine learning; and (b) the interpretability of the results. An overview of our research methods (more details in Sections 2 and 4) is as follows:

1. Represent the two classes of data (virus and worm signatures) in amino acids.
2. For each class, align the sequences using an appropriate method and substitution matrix (result is variable length samples in each class).
3. Represent the gaps with an amino acid.
4. Align all samples of all classes together (result is fixed length, and longer, sequences).
5. Apply machine learning (ML) techniques and compare with machine learning results using the original, unaligned data.
6. Extract rules.
7. Check the sequences from step 2 against naturally occurring proteins.
8. Check the consensuses from step 2 against naturally occurring proteins and give them biological realization.

9. Give the rules extracted at step 6 a biological realization and check against naturally occurring proteins.

Section 2 deals with steps 1-6 and Section 4 with steps 7-9.

## 2 SYSTEMS AND METHODS 1

Viruses and worms can be written in any programming language before being compiled. Source code will not be used in this paper. Instead, in line with malware signature detection, the signatures of viruses and worms, expressed in hexadecimal, are used in the experiments below. For instance, the first part of the *virus.1C.Tanga.a* computer virus signature is 8e5ef1aec9125; the first part of worm *Bat.Agent.bo* is fb56373bde388. The hexadecimal signatures for 60 viruses and 60 worms were downloaded from VX Heavens.

**Table 1: Three different representations (R1-R3) of hexadecimal code (bold) in amino acid (residue) alphabet**

| Hex | 1 | 2 | 3 | 4 | 5 | 6 | 7 | 8 | 9 |
|-----|---|---|---|---|---|---|---|---|---|
| R1  | A | C | D | E | F | G | H | I | K |
| R2  | S | R | Q | P | N | M | L | K | I |
| R3  | A | D | E | F | G | H | I | K | L |
| **Hex** | **0** | **a** | **b** | **c** | **d** | **e** | **f** | **-** | **-** |
| R1  | L | M | N | P | Q | R | S | Y | W |
| R2  | H | G | F | E | D | C | A | Y | W |
| R3  | M | N | P | Q | R | S | C | Y | W |

Three different representations of the hexadecimal code in the amino acid alphabet were tried for alignment purposes (Table 1). The first representation (R1, $2^{nd}$ and $6^{th}$ rows of Table 1) uses the same order of hexadecimal to amino acid residues and is the representation used previously (Chen *et al*., 2012a; Chen *et al*., 2012b). The second (R2, $3^{rd}$ and $7^{th}$) reverses this order and the third (R3, $4^{th}$ and $8^{th}$) uses a shift of one amino acid residue after the initial residue. This resulted in three datasets, each of 120 instances. In all cases 'W' is used to represent gaps introduced in the first alignment of worms and viruses separately, and 'Y' for gaps in the second alignment of worms and viruses jointly (details below).

Sequence alignment for multiple sequences typically performs a pair-wise alignment first and then uses information produced during pair-wise alignment to perform a multiple alignment. The same or different, or no, substitution matrices may be used during the two phases. These substitution matrices reflect the effects of pre-defined biological relationships between residues when alignments are formed during the first and second phases. If different amino acid representations are used for malware signatures, different alignments will be produced depending on the substitution matrix, the stage of alignment and the representation method. Substitutions can be weighted, however, as to their probability or likelihood given the sequences actually being aligned. The calculation of these weights is built into the alignment method used as default parameters. For instance, an alignment method that is two-phased may use the first phase to generate weights on residue substitutions based on complete pair-wise alignment, with these weights being used in a second phase global alignment for determining where to insert gaps. Default weighting parameters were used in all the experiments below.

In this paper, ClustalW (Thompson *et al*., 1994) and T-Coffee (Notredame et al., 2000) are used in four experiments for aligning the signatures represented in the three ways described in Table 1. ClustalW (as provided on the EBI site), in addition to using weightings (hence the 'W' suffix) on sequences so that similar sequences are down-weighted and dissimilar sequences are upweighted, allows one of four substitution methods, including no substitution matrix (called the 'identity' or 'unitary' matrix), to be used during both its initial pairwise alignment and its secondary multiple alignment. The first set of alignment experiments used ClustalW with default parameters except for the choice of substitution matrices, which were set to unitary matrices (1 for a positive self-match, 0 or negative for a mismatch) for both phases of alignment. This results in gaps being introduced as part of the alignment but not substitutions. The second set of alignment experiments with ClustalW switched on the substitution matrix BLOSUM (Henikoff and Henikoff, 1992) that reflects highly conserved regions of existing protein families. The third set of experiments with ClustalW used another substitution matrix, Gonnet (Gonnet *et al*., 1992). Very generally, Gonnet matrices represent evolutionary substitution and therefore distance information gained from analysing all protein sequences known in 1992, whereas BLOSUM matrices use conservation and similarity of function information within the sequences being analysed. BLOSUM can lead to alignments where amino acids are blocked in appearance in comparison to other alignment techniques that can disperse the amino acids more widely in the alignment. It is well known that different substitution matrices produce different results on the same set of sequences reflecting different purpose (Hara *et al*., 2010).

For the fourth and final alignment experiment, T-Coffee was used with BLOSUM as the substitution matrix. Default parameters are used to set gap penalties (based on an evolutionary model where gaps represent insertion or deletion mutations) after calculating a distance matrix from similarity scores for every possible pair of sequences in the first phase of alignment. T-Coffee then adds a position-specific scoring scheme by pooling information from the first phase for use in the second phase of multiple alignment. FASTA format (Lipman and Pearson, 1985) was used as the input method for all alignments.

The experimental method adopted is as follows:

*<u>Initial step:</u>* Convert the 60 worm signatures into three amino acid representation datasets using Table 1. Call these datasets WR1-WR3 (for 'worm representation 1' - 'worm representation 3'). Repeat for the virus signatures ('VR1' – 'VR3'). Each sample in the dataset is an artificial polypeptide sequence consisting of amino acid letters.

*<u>Step (a)</u>* Input all 60 WR1 worm polypeptide sequences into ClustalW using the unitary matrix to form an initial set of aligned worm sequences (SAWCIR1, for 'single aligned worm using ClustalW with Identity matrix and Representation 1'). Code gaps as 'W'. Repeat for WR2 and WR3 worms, resulting in SAWCIR2 and SAWCIR3. Store the three consensuses produced by the multiple alignment as WCCIR1-WCCIR3, for 'worm consensus using ClustalW with Identity matrix Representation 1' to 'worm consensus ClustalW with Identity matrix Representation 3'. We return to these consensus sequences in the second set of systems and methods (Section 4).

*<u>Step (b)</u>* Input all 60 VR1 virus polypeptide sequences into ClustalW (unitary matrix) to form an initial set of aligned virus sequences. Code gaps as 'W'. Repeat for VR2 and VR3 viruses. This results in three single aligned datasets SAVCWR1-SAVCWR3. Store the three consensuses produced by ClustalW (VCCWR1-VCCWR3) for later use in a second set of experiments (Sections 4 and 5 below).

***Step (c)*** Combine the two single aligned sets SAWCWR1 (worm) and SAVCWR1 (virus) into one set and input into ClustalW (unitary matrix) to form a second, doubly aligned set of polypeptide sequences. Code all gaps introduced at this (double alignment) stage as 'Y'. Repeat for R2 and R3 worm and virus singly aligned polypeptide sequences. This results in three datasets consisting of doubly multiply aligned worm and virus polypeptide sequences, with each data set containing sequences using R1, R2 and R3, respectively. Call these datasets DACIR1, DACIR2 and DACIR3 where 'DA' is 'doubly aligned' (i.e. combined, doubly aligned set), 'CI' is 'ClustalW with identity/unitary matrix' and R1-R3 are as described previously (three different representations using Table 1).

***Step (d)*** Repeat (a)-(c) but with the following changes:
    (i) switch on Gonnet within ClustalW for both phases of alignment;
    (ii) switch off Gonnet and switch on BLOSUM within ClustalW for both phases of alignment;
    (iii) use T-Coffee and switch on BLOSUM at the final stage of multiple alignment.

This results in 9 more doubly aligned datasets: DACGR1-3, where 'CG' is ClustalW with Gonnet; DACBR1-3, where 'CB' is ClustalW with BLOSUM; and DATBR1-3, where 'TB' is T-Coffee with BLOSUM switched on. The templates DAxyR1-DAxyz3 will be used as follows: x=C(lustal) or T(-Coffee); y= I(dentity) or G(onnet) or B(LOSUM). These twelve datasets will be input to five different symbolic machine learning techniques (details below) using their amino acid representations or, in the case of the perceptron, their numeric equivalents (see below). Table 2 (Appendix A) provides an overview of all 68 datasets generated by the initial step and steps (a)-(c) above.

***Step (e)*** For input to a neural network, convert all DAxyR1 – DAxyR3 datasets (twelve in total) from their polypeptide alphabetic representation into numeric using the replacements specified in Table 3. To take into account the arbitrary nature of the conversion of amino acid residues to numerical values, a hidden layer of 72 units is introduced. Previous work (Chen *et al*., 2012a) had shown that a 72-node hidden layer, in comparison to other architectures, was effective. The numeric conversion was also used successfully previously (Chen *et al*., 2012b) and is used again here. A two-layer network can also deal with the input values and their mapping to the output value in a non-linear way. Weka perceptrons were used to implement the neural networks, which have as many input nodes as amino acids in the fixed length, doubly aligned polypeptide sequences. For Weka, each residue position was given its own attribute name defined by its position in the doubly aligned sequence. Finally, for supervised machine learning, a '0' is affixed to each virus sample and a '1' to each worm sample.

*Step (f)* Test all three representations for generalizability and ability to withstand over-fitting across all machine learning techniques. The first condition of generalizability is implemented through a 66:34 training and testing regime, and the aim is to check how well the representations classify a relatively large proportion of unseen cases. The second condition of ability to withstand over-fitting is implemented through a 10-fold training and cross-validation regime, with 90% of samples used for training and 10% for cross-validation. Over-fitting can occur if the trained classifier has problems classifying a relatively small proportion of unseen cases. Report overall accuracy results for each of the machine learning techniques across both test conditions.

*Step (g)* Analyse all three representations for richness of knowledge extraction using PRISM (Cendrowska, 1987) available in Weka and analyse the results for possible viral and worm meta-signatures. Whereas previously J48 (a rule extractor with pruning) had been used (Chen *et al.*, 2012a), here we report on the application of PRISM (a modular rule extractor) to help compare the results. PRISM was used with no testing to maximise knowledge extraction (i.e. all samples were used for rule extraction).

**Table 3: Conversion of the 16 amino acid alphabet to numeric form between 0 to 1 for input to perceptrons. Y and W (two extra characters) represent the gaps introduced during alignment (see main text).**

| A | C | D | E | F | G | H | I | K |
|---|---|---|---|---|---|---|---|---|
| 0.1 | 0.15 | 0.2 | 0.25 | 0.3 | 0.35 | 0.4 | 0.45 | 0.5 |
| L | M | N | P | Q | R | S | Y | W |
| 0.55 | 0.6 | 0.65 | 0.7 | 0.75 | 0.8 | 0.85 | 0.9 | 0.95 |

*Step (h)* Evaluate the datasets with the best alignment and representation, as given by the machine learning results, as if they are real biological sequences to check for any added value to supplement the machine learning results. The methods adopted for this stage are described in more detail in Section 4 below.

Aligning the 60 virus and 60 worm polypeptide sequences separately (a-b above) allows the conserved regions of the virus and worm sequences across families within each class to be independently extracted. For instance, after alignment by T-Coffee, we have (for the first parts of three viral polypeptide sequences using R1):

```
FIIDIDNGLFDSRPLEEFKGALEGEI...
GE-----SQMPSIDMPQF---PGLPS...
---------ILHSPMHQFRF-PRSQR...
                : :*
```

which shows that only F is aligned across all three sequences ('*'), and M and Q across two sequences (":"). The gaps '-' introduced at this stage are coded 'W'. When these aligned virus sequences are themselves aligned with their worm sequence counterpart representations, the gaps introduced are coded 'Y' (step (c) above). Y and W have their own representation for perceptron input (Table 2).

Four symbolic machine learning algorithms were used in addition to the (numeric) perceptron: Naïve Bayes, J48, LAD Tree and OneR. Naïve Bayes is a simple version of Bayesian classifiers that only looks at the relationship between a particular feature and classification, i.e. no attention is paid to combinations of features. J48, as noted earlier, is an implementation of ID3 and C4.5 tree induction algorithms and uses information-theoretic formulae (Quinlan, 1993). The LAD tree algorithm is an approach using least absolute deviation (LAD) error to obtain regression trees that can handle discrete variables through recursive partitioning (Breiman *et al*., 1983). OneR is a single rule (one rule) classification algorithm that produces a tree of only one level, where the rule is a set of attributes associated with their majority class (Holte, 1993). For reporting the test results, the following standard formula for accuracy using numbers of samples in each category is adopted (virus is negative (N), worm is positive (P), true is T and false is F)*: accuracy = (TP+TN)/(TP+TN+FP+FN).* No attempt is made to distinguish sensitivity from specificity, since it is the overall effect of the representation on accuracy that is being tested.

# 3   RESULTS 1

Benchmark classification accuracy figures were first calculated for the non-aligned (original) sequences (VR1+WR1, VR2+WR2, VR3+WR3), using 66:34 training and testing ratio, and 10 fold cross validation using 90:10:

Naïve Bayes: 26% (66:34), 43% (90:10)

J48: 54%, 54%

LAD Tree: 51%, 63%

OneR: 49%, 58%

Perceptron: 39%, 48%

The overall benchmark accuracy average was 44% for the 66:34 ratio and 53% for the 90:10 ten-fold cross validation (49% overall benchmark average) for the non-aligned sequences. The accuracy remains the same for all three modes of representation, indicating (as expected) that these machine learning techniques treat the representations as context-free. These benchmark accuracy figures indicate that the 120 signatures, in their original hex and amino acid forms, are problematic from a machine learning perspective, i.e. not much better than tossing a coin. The multiple alignments of worms and viruses (steps (a)-(c) above) resulted in fixed length sequences varying in length as described in Table 4.

**Table 4: The effect on length of viral and worm signatures by representation (R), method (ClustalW or T-Coffee) and substution matrix (Identity, Gonnet, BLOSUM)**

|  | ClustalW with ID (xy=CI) | ClustalW with Gonnet (xy=CG) | ClustalW with BLOSUM (xy=CB) | T-coffee with BLOSUM (xy=TB) |
|---|---|---|---|---|
| **VIRUS–R1** | 121 | 91 | 81 | 223 |
| **VIRUS-R2** | 116 | 85 | 79 | 193 |
| **VIRUS-R3** | 140 | 85 | 79 | 212 |
| **WORM-R1** | 129 | 90 | 87 | 185 |
| **WORM-R2** | 121 | 87 | 87 | 202 |
| **WORM-R3** | 116 | 89 | 84 | 233 |
| **DAxyR1** | 494 | **171** | 103 | 981 |
| **DAxyR2** | 440 | **112** | 106 | 1101 |
| **DAxyR3** | 462 | 128 | 86 | 1284 |

There were significant differences in length by the four alignment methods (i.e. six pairwise comparison of the columns of Table 4, paired samples T-test, −424.68≤t≤149.68, df=8, p≤0.05). There was no difference between lengths by alignment method by representation (i.e. repeating the above paired samples T test for representation 1 only, for representation 2 only and then representation 3 only, 18 pairwise tests altogether).

The overall results for all four symbolic algorithms and the perceptron, under the two test conditions (66-44 training testing, and 10-fold cross-validation), are presented in Table 5. A fixed learning rate of 0.2 and momentum of 0.1 were used for all 12 perceptrons (3 representation methods x 4 alignment methods) for 200 epochs (the same as for the benchmarked perceptron results). One sample T tests comparing the results of each of the five ML techniques using aligned and combined virus and worm sequences showed significant differences (improvements) over the benchmark, non-aligned results (i.e. columns a-m tested against the original unaligned benchmark figures above column index in Table 5): 5.15≤t≤11.89, p≤0.001, df=11, 13 tests. There was only one significant difference between representations in terms of classification accuracy: J48 under the test condition 90/10 cross validation (column d of Table 5) between representations 1 and 3 (independent samples T test, t=2.62, df=6, p≤0.05). There were some significant differences between alignment methods and specific ML classification accuracies (paired samples T test, df=4 in all cases, p≤0.05, row and column numbers as per indexes in Table 5):

    Between CG and CB (rows 2,3,7,8,12,13): perceptron 66/34 (column i), t=4.3.
    Between CG and TB (2,4,7,9,12,14): Naïve Bayes 66/34 (a), t=4.84; Naïve Bayes 90/10 (b), t=7.2; J48 66/34 (c), t=3.28.
    Between CB and TB (3,4,8,9,13,14): Naïve Bayes 66/34 (a), t=4.35; J48 66/34 (c), t=3.04; OneR 90/10 (h), t=3.0.

**Table 5: Accuracy figures for R1, R2 and R3 representations across five machine learning algorithms after four different methods of alignment CI, CG, CB and TB. '66/34' and '90/10' refer to the training, testing and cross-validation regimes used.**

|  |  | Naïve Bayes | | J48 | | LAD Tree | | OneR | | Perceptron | | Average | | Overall average |
|---|---|---|---|---|---|---|---|---|---|---|---|---|---|---|
|  |  | 66/34 | 90/10 | 66/34 | 90/10 | 66/34 | 90/10 | 66/34 | 90/10 | 66/34 | 90/10 | 66/34 | 90/10 |  |
|  | Original unaligned | 0.26 | 0.43 | 0.54 | 0.54 | 0.51 | 0.63 | 0.49 | 0.58 | 0.39 | 0.48 | 0.44 | 0.53 | 0.49 |
|  | column index | a | b | c | d | e | f | g | h | i | j | k | l | m |
| row index | **R1** | | | | | | | | | | | | | |
| 1 | CI | 0.49 | 0.60 | 0.49 | 0.77 | 0.66 | 0.72 | 0.56 | 0.59 | 0.51 | 0.54 | 0.54 | 0.64 | 0.59 |
| 2 | CG | 0.95 | 0.98 | 0.88 | 0.83 | 0.93 | 0.91 | 0.78 | 0.81 | 0.95 | 0.97 | 0.90 | 0.90 | 0.90 |
| 3 | CB | 0.90 | 0.91 | 0.90 | 0.82 | 0.90 | 0.85 | 0.81 | 0.85 | 0.83 | 0.79 | 0.87 | 0.84 | 0.86 |
| 4 | TB | 0.71 | 0.90 | 0.73 | 0.82 | 0.93 | 0.91 | 0.78 | 0.80 | 0.85 | 0.95 | 0.80 | 0.88 | 0.84 |
| 5 | *Average R1* | 0.76 | 0.85 | 0.75 | 0.81 | 0.85 | 0.85 | 0.73 | 0.76 | 0.79 | 0.81 | 0.78 | 0.82 | 0.80 |
|  | **R2** | | | | | | | | | | | | | |
| 6 | CI | 0.63 | 0.64 | 0.73 | 0.65 | 0.71 | 0.72 | 0.51 | 0.64 | 0.71 | 0.58 | 0.66 | 0.65 | 0.65 |
| 7 | CG | 0.95 | 0.97 | 0.98 | 0.88 | 1.00 | 0.98 | 0.88 | 0.97 | 0.98 | 0.96 | 0.96 | 0.95 | 0.95 |
| 8 | CB | 1.00 | 0.98 | 1.00 | 0.95 | 1.00 | 0.99 | 0.98 | 0.98 | 0.88 | 0.89 | 0.97 | 0.96 | 0.96 |
| 9 | TB | 0.44 | 0.78 | 0.85 | 0.78 | 0.73 | 0.71 | 0.61 | 0.56 | 0.85 | 0.86 | 0.70 | 0.74 | 0.72 |
| 10 | *Average R2* | 0.76 | 0.84 | 0.89 | 0.82 | 0.86 | 0.85 | 0.74 | 0.79 | 0.85 | 0.82 | 0.82 | 0.82 | 0.82 |
|  | **R3** | | | | | | | | | | | | | |
| 11 | CI | 0.98 | 0.90 | 1.00 | 0.98 | 0.98 | 0.92 | 0.90 | 0.87 | 0.95 | 0.93 | 0.96 | 0.92 | 0.94 |
| 12 | CG | 1.00 | 0.99 | 0.95 | 0.96 | 0.98 | 0.99 | 0.95 | 0.97 | 0.98 | 0.98 | 0.97 | 0.98 | 0.97 |
| 13 | CB | 0.93 | 0.98 | 0.90 | 0.97 | 0.98 | 0.99 | 0.98 | 0.99 | 0.90 | 0.96 | 0.94 | 0.98 | 0.96 |
| 14 | TB | 0.59 | 0.87 | 0.78 | 0.80 | 0.76 | 0.82 | 0.76 | 0.71 | 1.00 | 0.94 | 0.78 | 0.83 | 0.80 |
| 15 | *Average R3* | 0.87 | 0.94 | 0.91 | 0.93 | 0.92 | 0.93 | 0.90 | 0.89 | 0.96 | 0.95 | 0.91 | 0.93 | 0.92 |
|  |  | | | | | | | | | | | | | |
| 16 | *Average ML method* | 0.80 | 0.88 | 0.85 | 0.85 | 0.88 | 0.88 | 0.79 | 0.81 | 0.87 | 0.86 | 0.84 | 0.86 | 0.85 |

There were also significant differences between some alignment methods and average classification accuracies in Table 5 (independent samples T Test, df=4 in all cases, $p \leq 0.05$):

Between CG and TB (rows 2,4,7,9,12,14): Average 66/34 (k), t=4.8; Average 90/10 (l), t=2.69; Overall average (m), t=3.74.

Between CB and TB (3,4,8,9,13,14): Average 66/34 (k), t=3.92; Overall average (m), t=2.9.

In other words, TB appeared to have most (negative) effect on average classification accuracies in comparison to CG and CB. Analysis of variance (ANOVA) showed no difference in the variances of the individual ML technique accuracies by representation and only three significant differences ($p \leq 0.05$, df=11) by alignment: Naïve Bayes 66/34, F=4.50; Naïve Bayes 90/10, F=5.53; and OneR 90/10, F=4.24. Post-hoc tests showed that, in the case of the first, TB had significantly different negative means in comparison to CG and CB only; in the case of the second, CI had a significantly different negative mean in comparison to CG and CB only; and similarly for the third case. In other words, in no case were all the means within an ML technique significantly different from each other by alignment. Although there was no overall significant difference between alignment methods as a group and combined overall average, post-hoc ANOVA analysis showed that CI had a significantly lower mean than CG and CB ($p \leq 0.05$). Table 6 provides the average accuracies by alignment method to make these differences clearer. There were a number of other ANOVA differences significant in the range $0.051 \leq p \leq 0.1$ (and visually identifiable in Table 6) and not reported here. In other words, CI was on the whole the worst performing alignment method. This may seem surprising, given that CI is the one alignment technique that does not use biologically plausible substitution matrices and only inserts gaps to help identical amino acids in different sequences to align in columns. However, identity matrices will work best when there is significant commonality between sequences. Their relatively poor performance here indicates that the sequences do not share much commonality.

Table 6: Averaged alignment accuracy figures of Table 5. The underlined figures show the significant differences found through ANOVA, with figures in bold indicating the entries that these underlined figures differed significantly from.

| alignment | NB 66/34 | NB 90/10 | J48 66/34 | J48 90/10 | LAD 66/34 | LAD 90/10 | OneR 66/34 | OneR 90/10 | P 66/34 | P 90/10 | Average 66/34 | Average 90/10 | Combined overall average |
|---|---|---|---|---|---|---|---|---|---|---|---|---|---|
| CI | 0.70 | **_0.71_** | **_0.74_** | 0.80 | 0.78 | 0.79 | 0.66 | **_0.70_** | 0.72 | 0.68 | 0.72 | 0.74 | **_0.73_** |
| CG | **0.97** | **0.98** | **0.94** | 0.89 | 0.97 | 0.96 | 0.87 | **0.92** | 0.97 | 0.97 | 0.94 | 0.94 | **0.94** |
| CB | **0.94** | **0.96** | **0.93** | 0.91 | 0.96 | 0.94 | 0.92 | **0.94** | 0.87 | 0.88 | 0.93 | 0.93 | **0.93** |
| TB | _0.58_ | 0.85 | 0.79 | 0.80 | 0.81 | 0.81 | 0.72 | 0.69 | 0.90 | 0.92 | 0.76 | 0.82 | 0.79 |

Overall, R3 performed best (0.92 overall accuracy, row 15, column m of Table 5) across all five ML techniques under both test conditions, irrespective of alignment method. ClustalW with Gonnet was the best alignment method (0.94) across all ML techniques, irrespective of representation (Table 6). All accuracy results using aligned data were individually better than the benchmark accuracy for the non-aligned datasets. The overall accuracy of 85% (bottom right of Table 5) is a major improvement on the 49% accuracy for the benchmarked (non-aligned) data, demonstrating that the use of sequence alignment techniques can serve a useful ML purpose.

To complete the first set of results, the best sequences of the three representations, as given by the overall accuracy figures in Table 5, were input to the rule extractor, PRISM. R1 worked best with ClustalW using Gonnet (0.9, row 2), R2 with ClustalW using BLOSUM (0.96, row 8) and R3 with ClustalW using Gonnet (0.97, row 12). All 120 doubly aligned sequences were used for rule extraction (i.e. no training and testing) to maximize knowledge extraction. For R1 CG sequences, the following rules were found (all W and Y gap representations are excluded; 'pos' refers to 'position' in the sequence):

> Virus: If pos36 = A, pos21 = D, pos28 = E, pos53 = A, pos20 = N, pos5 = A, pos30 = L, pos32 = A, pos36 = P.
> Worm: If pos72 = L, pos73 = P, 51 = H, pos59 = S, pos70 = R, pos73 = D, pos46 = R, pos72 = M, pos71 = S, pos44 = I, pos45 = L, pos70 = G, pos10 = L, pos41 = C, pos45 = D, pos54 = L.

Rewritten as a left to right amino acid sequence using the positional information, these rules produce the strings ..A..ND..E..L..A..[AP]..A.. for virus and ..L..C..I[LD]R..H..L..S..[GR]S[LM][DP].. for worm, where '..' stands for 'any number of any amino acid' and square parentheses indicated alternatives. Converting these amino acids sequences back to their R1 hexadecimal equivalents using Table 1 and removing the gaps produce the meta-signatures '1b3401[1c]1' for virus and '028[03]e70f[6e]f[0a][3c]' for worm.

For R2 CB, rules with only gaps except for 'worm if pos21=R' were returned. For R3 CG, the following rules were found by PRISM:

> Virus: If pos65 = F, pos12 = A, pos13 = A.
> Worm: If pos124 = M, pos122 = A, pos124 = H, pos119 = N, pos55 = M, pos88 = M, pos11 = A, pos33 = I, pos55 = L.

Rewritten as amino acid, left to right sequences, these rules produce ..AA..F.. for virus and ..A..I..[LM]..M..N..A..[HM].. for worm, which in turn produce the hex meta-signatures '114' for virus and '17[90]0a1[60]' for worm.

In other words, these meta-signatures represent those parts of the virus and worm signatures that are conserved between variants and families of virus and worm signatures as well as those parts that distinguish viral signatures from worm signatures, using R1 and R3 as the representation methods, ClustalW as the alignment method and Gonnet as the substitution matrix used for both phases of alignment within ClustalW.

## 4 SYSTEMS AND METHODS 2

The results above indicate that ClustalW with Gonnet (CG) produces the most accurate classification results (0.94 in Table 6) as well as most interpretable and useful meta-signatures for data mining purposes, especially when R1 is used as the representation method. The meta-signatures produced by PRISM using CG, namely, '1b3401[1c]1' for R1 virus and '028[03]e70f[6e]f[0a][3c]' for R1 worm, and '114' for R3 virus and '17[90]0a1[60]' for R3 worm, should ideally be interpreted by tracing back to the op codes in the original source code from which the hexadecimal signatures were initially derived. But the malware source code is not available in this case (due to security concerns regarding the public dissemination of malware code) and may not be available for other datasets where there has been a non-reversible transformation or conversion from the original data to that used for data mining. The aim of the second set of experiments is to determine the feasibility of interpreting these meta-signatures by reference to the biological domain.

The linear representation of sequences represents the primary structure of the protein. The secondary structure of an amino acid sequence provides an indication of how segments can form a 3D structure through local interactions. The basic secondary structures are 'helix', 'coil' ; 'sheet' and 'beta turn', and can be calculated from the hydrogen bonds between amino acids and carboxyl groups. Secondary structure gives rise to tertiary structure, which is the 3D representation of the entire sequence determined by atomic coordinates. Many sources now exist that describe the relationships between these three structures and the problems associated with computational prediction of secondary and tertiary structure from the primary sequence.[1] Finally, there is quaternary structure, which is 3D representation of a complex protein structure consisting of two or more functionally distinct subsequences.

One way to check for interpretability is to see whether a particular representation, after alignment, has biological plausibility in terms of matches with real protein sequences. That is, the 'most plausible representation' can be hypothesized to be the one that maps signatures of a class, after initial alignment to identify commonalities across families within that class, to the most biological analogues, whereas the' least plausible representation' is one that maps to few biological analogues. Using a representation that has more biological plausibility may, in turn, be reflected in improved alignment due to the use of biologically based substitution matrices that reflect known frequencies of mutations and conserved subsequences of residues in existing protein sequences.   The first set of experiments in this section takes each of the 120 sequences in the singly aligned datasets using all three representations and entering them, one by one, into the PRINTS database as queries to see which of the existing known proteins are returned as the closest match between the artificial, aligned polypeptide sequence and naturally occurring polypeptide sequences. PRINTS (Attwood *et al*. 1994; Attwood *et al*., 2012) consists of conserved motifs (called 'fingerprints') found in already aligned proteins with known function and structure. Newly aligned sequences can be matched against these fingerprints to tentatively assign these sequences to known families of protein and hence predict the function and structure of these new sequences. The number of hits in the first experiment will help us determine the most plausible representation as given by the highest number of hits. To provide a benchmark, the original, unaligned sequences in all three representations were also input to PRINTS to see if alignment produced better results. These experiments investigate the three primary structure representations (R1-R3) produced by the four alignment methods in more detail.

---

[1] See http://en.wikipedia.org/wiki/Tertiary_structure#Tertiary_structure for an initial set of references.

Since ClustalW with Gonnet introduces biological plausibility through the use of Gonnet substitution matrices, it would be useful to adopt 'protein semantics' as an approach to interpretability of results. The second set of experiments below visualizes the secondary and tertiary structure of the consensuses produced by ClustalW using Gonnet on R1, R2 and R3 to identify any biological commonalities across representation types that have been introduced by the biological substitution matrices. These consensuses were produced during the first phase of alignment (i.e. when virus signatures were aligned separately from worm signatures - steps (a) and (b) in Section 2)). These visual realizations may contain useful structural information to supplement the machine learning results. For the second set of experiments below, the six consensuses (WCCGR1-3, VCCGR1-3) were analyzed and only amino acids that achieved a threshold 15% appearance in the consensus were kept for realization. This represents a minimum of 9 appearances of that amino acid in a specific position in the 60 signatures after alignment by ClustalW using Gonnet.  Where two or more amino acids achieved this threshold for a particular position in the alignment, all such amino acids were listed in the abbreviated consensus. These abbreviated consensuses were entered into the *ab initio* 3D protein modeler QUARK (Xu and Zhang, 2012) for secondary structure analysis and tertiary structure realization. Such *ab initio* modelers use the atomic and molecular properties of individual amino acids, irrespective of whether the sequence actually exists as a protein.

Finally, the meta-signatures produced by PRISM and using ClustalW with Gonnet can be given biological meaning by calculating their secondary and tertiary structure predictions, and then combining them into a complex multiple protein structure. Such a quaternary structure is the arrangement of multiple folded proteins, where conformational changes can lead to re-orientation of subunits and allostery – the regulation of the complex multiple protein through binding effector molecules on several binding sites rather than just one.[2] The quaternary structure of the meta-signatures provides 'super-representational' structure for both the virus and worm meta-signatures with the binding sites differing according to the representation used. These signatures can be checked against known signatures of protein families using PRINTS available at EBI.

In summary, the experimental method for the second set of experiments is as follows:
***Step (i):*** Take the six non-aligned datasets VR1-VR3 and WR1-WR3 and find the number of hits against known proteins using each sequence as a fingerprint. Repeat for the singly aligned versions of these datasets, i.e. SAVCIR1-R3, SAWCIR1-R3, SAVCGR1-R3, SAWCGR1-R3, SAVCBR1-R3, SAWCBR1-R3, SAVTBR1-R3, SAWTBR1-R3 (see Table 2, Appendix A for a description of these datasets). Compare the results.
***Step (j):*** Take the six consensuses produced by ClustalW with Gonnet using all three representations (i.e. VCCGR1-R3 and WCCGR1-R3), apply a 15% appearance threshold to each member of the consensus, and input the abbreviated consensuses to QUARK for secondary and tertiary structure prediction and realization. Report any interesting features.
***Step k:*** Combine the R1 and R3 meta-signatures produced by PRISM and using ClustalW with Gonnet into one long sequence and calculate the quaternary structure of these combined meta-signatures. Check for any interesting similarities and/or differences both visually and using PRINTS.

---

[2] See http://en.wikipedia.org/wiki/Protein_quaternary_structure for an initial introduction to quaternary structure and allostery.

# 5 RESULTS 2

For the first experiment, Table 7 describes the number of hits by sequence and in total for each of the alignment methods, including non-alignment as a benchmark. As can be seen, TB provides the least biological plausibility (only one hit) and CG the most in terms of total hits against proteins (368) and the most number of different proteins found (349). CI was less plausible than the benchmarked non-aligned sequences in terms of hits on all three measures.

Table 7: The number of hits against existing proteins using PRINTS before and after the first alignment. The rows represent the 60 signatures represented using R1-R3. The columns represent the alignment method. 'SC' provides the total number of sequences out of 60 in that particular dataset that were matched against existing proteins; 'TC' provides the total number of proteins matched by all sequences in that dataset; 'PC' provides the total number of different proteins matched by all sequences in that dataset.

|  | Non-aligned | | | CI | | | CG | | | CB | | | TB | | |
|---|---|---|---|---|---|---|---|---|---|---|---|---|---|---|---|
|  | SC | TC | PC | SC | TC | PC | SC | TC | PC | SC | TC | PC | SC | TC | PC |
| R1-virus | 15 | 39 | 39 | 14 | 51 | 51 | 24 | 65 | 63 | 23 | 68 | 65 | 0 | 0 | 0 |
| R1-worm | 15 | 37 | 37 | 8 | 22 | 22 | 12 | 32 | 32 | 17 | 47 | 47 | 0 | 0 | 0 |
| R2-virus | 18 | 42 | 42 | 11 | 22 | 22 | 29 | 93 | 85 | 22 | 64 | 64 | 0 | 0 | 0 |
| R2-worm | 12 | 29 | 29 | 11 | 32 | 32 | 16 | 41 | 41 | 18 | 46 | 46 | 0 | 0 | 0 |
| R3-virus | 9 | 18 | 18 | 4 | 9 | 9 | 22 | 83 | 82 | 17 | 48 | 48 | 1 | 1 | 1 |
| R3-worm | 17 | 32 | 32 | 9 | 19 | 19 | 16 | 54 | 46 | 12 | 46 | 44 | 0 | 0 | 0 |
| Totals | 86 | 197 | 197 | 57 | 155 | 155 | 119 | 368 | 349 | 109 | 319 | 314 | 1 | 1 | 1 |

CG showed a small but interesting reduction in the number of different proteins found in comparison to total proteins, indicating some common hits against the same protein by different sequences belonging to R1 virus (from 65 to 63), R2 virus (from 93 to 85), R3 virus (from 83 to 82) and R3 worm (from 54 to 46). Similarly, CB showed some small reductions for R1 virus (from 68 to 65) and R3 worm (from 46 to 44). These common hits were examined in more detail.

For CB R1 virus, the common proteins were (using PDB fingerprint ID codes) 1EZV (twice), 1KYO (twice) and 1KB9 (twice). 1EZV,1KYO and 1KYO all refer to ubiquinol-cytochrome reductase complex core proteins found in *S. cerevisiae* (bc1 complex) and play a key part in energy conversion of the respiratory and photosynthetic electron transfer chains (Hunte *et al*., 2000). For CB R3 worm, the common proteins were 1COV (twice) and 1JEW (twice), both of which are parts of the coxsackiervirus. 1COV is part of the coat (Muckelbauer *et al*., 1995) and 1JEW is a receptor (He *at al*., 2001). Coxsackievirus is an Entovirus, one of the most common human pathogens.[3]

For both the CG R1 virus and CG R1 worm, the common proteins were 1PG4 (twice for R1 virus; five times for R1 worm) and 1PG3 (twice for R1 virus; five times for R1 worm). 1PG4 and 1PG3 are both acetyl CoA synthetases from *Salmonella enterica* (Gulick *et al*., 2003). Synthetases (or ligases) catalyze the joining together of two molecules.[4] *S.enterica* is one of the most common causes of illness caused by food infection.[5]

For the second experiment, the following six abbreviated signature consensuses (WCCGR1-3 and VCCGR1-3 of Table 5) are formatted as follows: the first line is the representation and type identifier, with original consensus length as given in Table 4 followed by the abbreviated consensus length using a 15% residue commonality threshold; the second line is the abbreviated consensus; and third line is the secondary structure QUARK predictions (C = coil; H = helix; E = sheet; T = beta turn). Options for a specific position achieving the 15% threshold are in parentheses.

R1 virus consensus (original length 91, abbreviated length 32) :
```
CCCKQCAKF[AMP]IKRCIGPCNGP[FM]CD[AG]LKC
CHHHHHHHH[HHH]HHHTTTTCCCC[EE]ET[TT]TCC
```

R1 worm consensus (90, 29):
```
NFP[AL]NMCRMRGCHLPKCME[LM]QSHICRN
CCH[HH]HHHHHTTTTHHHHH[HH]HHHHCC
```

R2 virus consensus (85, 30):
```
CRSAGFGCSSNLCRLERCS[CI]LCRRRIECI
TTTTCCCCTTTTEEHCCCC[EE]EEEEEEECC
```

---

R2 worm consensus (87,28):

CFMFRGGCRCEGKCGCFK[DG]PFDCGPRL

CEETTTTEETTTTTTTCT[TT]TTTTTTTC

R3 virus consensus (85, 39):

CCQLFKAFAQCCGRCGKICQLGCLRI[IL]PEQCDLM[DP]IL

CHHHHHHHHHHHHHHHHHHHHCCEEE[EC]HHHCTTT[TT]TT

R3 worm consensus (89,34):

GLCPHDMPPHLIDC[EQ]CE[FL]FDCG[FIM][DP]CPCSC

CCTTTTCTTTTCCH[HH]TT[TT]TTTT[TTT][TT]TTCCC

These consensus sequences produced the 3D tertiary structure protein models in Figures 1-6, with options listed (arrows represent the primary and secondary sequence order of amino acids):

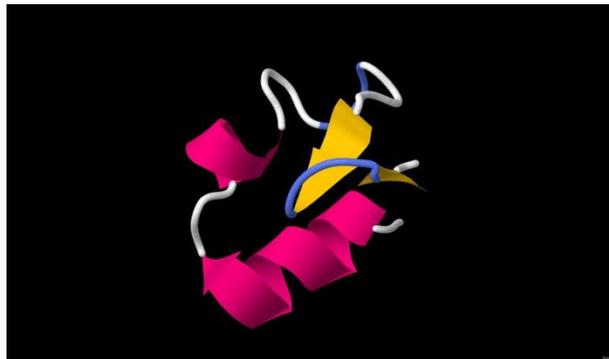
**Figure 1: Jmol visualization of R1 CG virus consensus**

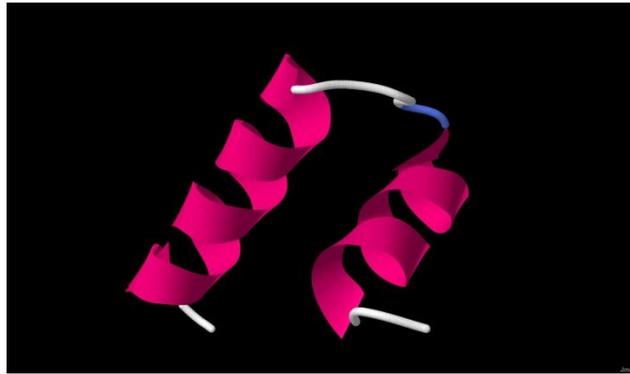
**Figure 2: Jmol visualization of R1 CG worm consensus**

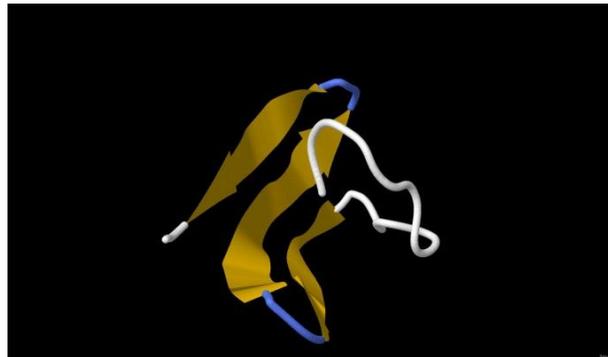
**Figure 3: Jmol visualization of R2 CG virus consensus**

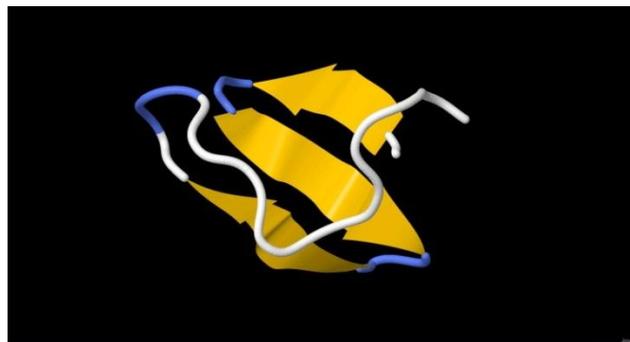
**Figure 4: Jmol visualization of R2 CG worm consensus**
2323

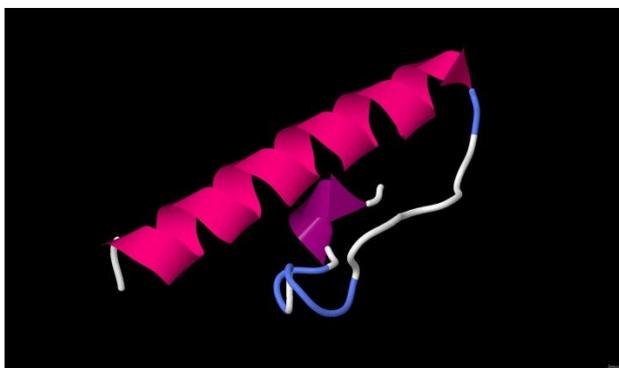
**Figure 5: Jmol visualization of R3 CG virus consensus**

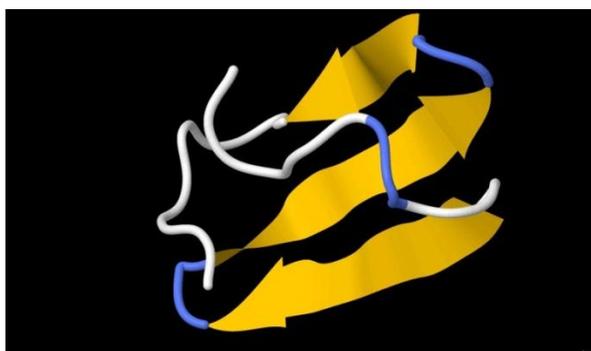
**Figure 6: Jmol visualization of R3 CG worm consensus**

Structurally and visually, R1 virus and R1 worm abbreviated consensuses (Figures 1 and 2) appear to be distinguished by an extra beta-pleated sheet in R1 virus (yellow signifies sheets that can be formed by a number of amino acid beta turns). R2 virus and R2 worm (Figures 3 and 4) look indistinguishable once their sheets are formed. Although the consensus was formed with ClustalW using Gonnet, this may also reflect in protein semantic terms the lack of informative rules found by PRISM using R2 with ClustalW and BLOSUM. R3 virus and R3 worm are clearly distinguishable through the presence of helices in R3 virus and beta-pleated sheets in R3 worm (Figures 5 and 6). White and blue strings signify random coils and loops. The arrows provide an indication of left-to-right direction along the primary structure.

For the third, and final, experiment, the meta signatures formed by PRISM using CG R1 virus and worm signatures (ANDELA[AP]A and LCI[LD]RHLS[GR]S[LM][DP], respectively) and R3 virus and worm signatures (AAF and AI[LM]MNA[HM], respectively) were conjoined and input to Quark (length 37). There are two ways to conjoin the four meta-signatures: R1 virus + R1 worm + R3 virus + R3 worm; and R1 virus + R3 virus + R1 worm + R3 worm. The former (R1V+R1W+R3V+R3W) keeps the representations together but splits the virus and worm meta-signatures, whereas the latter (R1V+R3V+R1W+R3W) keeps the virus and worm meta-signatures together but splits the representations. The first method produces:

ANDELA[AP]ALCI[LD]RHLS[GR]S[LM][DP]AAFAI[LM]MNA[HM]

and the second:

ANDELA[AP]AAAFLCI[LD]RHLS[GR]S[LM][DP]AI[LM]MNA[HM]

These two meta-signature sequences produced the quaternary structure shown in Figures 7 and 8. For R1V+R1W+R3V+R3W two small helices (first part of structure) are followed by a third, larger helix (second part), whereas for R1V+R3V+R1W+R3W only two large helices are formed. The first helix can be interpreted as R1V+R3V and the second as R1W+R3W. Random coils and loops occur in other parts.

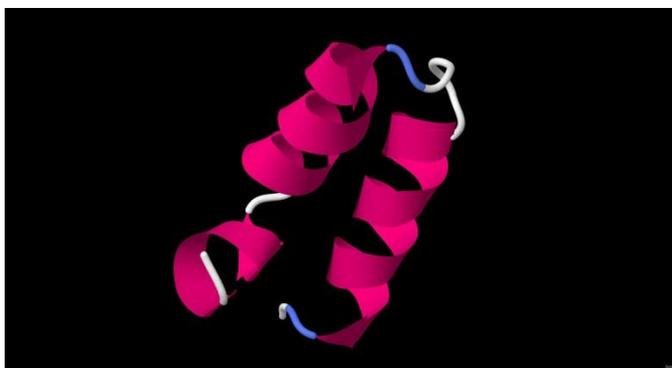

**Figure 7: JMol visualization of R1V+R3V+R1W+R3W PRISM rules (meta-signatures)**

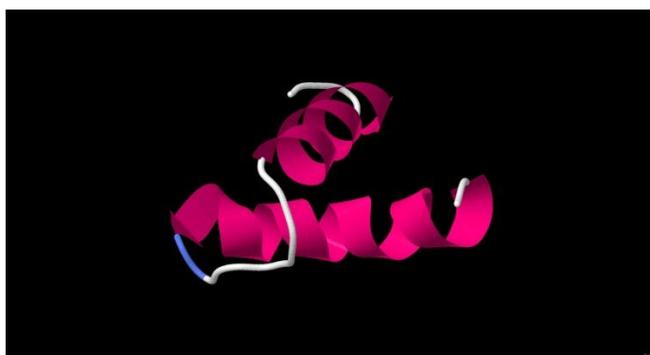

**Figure 2: JMol visualization of R1V+R1W+R3V+R3W PRISM rules (meta-signatures)**

When R1V+R1W+RF3V+R3W was checked using PRINTS, three matches were found. With 58% identity was Beta-lactoglobulin signature, a sequence containing a six-element fingerprint for a family of lipocalins that was itself formed from aligning five sequences. Lipocalins are pro-teins that extra-cellular proteins believed to be important for nutrient transport and control of cell regulation (Flower, 1996). With 55% similarity was orphan nuclear receptor NOR1 superfamily signature. This superfamily of nuclear hormone receptors consist of transcription regulators that mediate the transcription of target genes for a wide variety of functions, including embryonic development, cell differentiation and homeostasis. Nuclear receptors have been implicated in disease such as cancer and diabetes, and those that have not yet been clearly identified with specific targets are termed 'orphan'. In other words, NOR1 stands for a family of related nuclear receptors, the targets of which are not yet fully understood or known. With 50% similarity was B1 bradykinin receptor signature representing a family of G protein-coupled receptors (GPCRs) involved in, among other functions, activation of sensory fibres, cytokine release (Watson and Arkinstall, 1994).

When the conjoined meta-signature sequence R1V+R3V+R1W+R3W was checked, several matches were found. With 61% similarity was Melanocortin receptor family signature, another member of the GPCR family. Next with 60% similarity and 57% similarity were Beta-lactoglobulin signature (see above) and orphan nuclear receptor NOR1 superfamily signature (see above). With 54% similarity was another member of the GPCR super-family, Beta-3 adrenergic receptor signature. With 50% similarity and 46% were two more members of the GPCR super-family, B1 bradykinin receptor signature (see above) and Glycoprotein hormone receptor signature. With 31% similarity was Integrin A, a cell adhesion molecule family that is a signature for eight subunits found in a variety of different organisms including mouse, human and chick. Integrins are receptors on the surface of cells that have, as one of their functions, the transfer of information from outside the cell into the cell (Hynes, 1987).

# 6   CONCLUSION AND FURTHER WORK

The results of the first set of experiments show the feasibility of a bio-inspired data mining method for dealing with problematic machine learning data. The method of converting malware hexadecimal signatures to amino acid representation has been clearly demonstrated to affect learning and therefore the signatures extracted (Table 5). Given that there are $^{20}P_{16}$ ways to undertake the hex-residue conversion, there is clearly much more work required to identify

'optimal' coding schemes taking into account the purposes of data mining. The disadvantage with R1 is that it appears to produce the longest doubly aligned sequences of all three representations (Table 4). But the machine learning results indicate that longer sequences lead to greater accuracy and that the rules derived for generating malware meta-signatures are relatively information rich in comparison to shorter alignments. All accuracy figures on aligned sequences were better than the original benchmark accuracy figures for the non-aligned sequences, indicating that, for this dataset, alignment leads to major improvements in classification accuracy (Table 5). However, amino acid representation can only code for a maximum of 20 different attribute values, including gaps. Continuous data or categorical data consisting of more than 18 values (if two separate amino acids are used to represent gaps) may need to be binned to a maximum of 18 or 19 different values to allow sequence alignment using amino acids to be used. Nevertheless, the use of ClustalW with Gonnet as the substitution matrix for aligning problematic sequences has been shown to result in greater machine learning accuracy and this method may be of use with other problematic data.

More work is required to determine the trade-off between doubly aligned sequence length and richness or usefulness of signatures extracted. In particular, further work is required to determine how predefined substitution matrices should be used, given that the 'biology-free' unitary or identity matrix produces the worst machine learning results (Table 6). Converting the hexadecimal code of viruses and worms to amino acids and then rational numbers between 0 to 1 for input to neural networks has also been shown to be effective, provided that the viral and worm sequences are doubly aligned appropriately (ClustalW with Gonnet, 0.97 accuracy, Table 6). Overall, the first research question as to whether it is possible to use sequence techniques from bioinformatics to help distinguish between computer virus and worm signatures has been answered positively in this particular case.

The results of the second set of experiments show that, once samples are represented as biosequences, protein modeling may offer new ways of interpreting the data and rules extracted by machine learning. If a 'discriminatory' representation is required, where the samples are the most dissimilar to naturally occurring proteins, R3 is the best before alignment and any representation using T-Coffee with BLOSUM after alignment (Table 7). If a 'generalizable' representation is needed, alignment using ClustalW with Gonnet gives the most hits (Table 7) for all three representation methods in comparison to the other alignment methods. Both ClustalW with Gonnet and ClustalW with BLOSUM returned hits of aligned sequences against a small number of naturally occurring 'dangerous' proteins, i.e. proteins associated with infection. This could be coincidence or it could show some implicit relationship between the substitution matrices used for alignment and their construction from bacterial and viral databases in the past. Further work is required to determine whether more can be made of such implicit relationships or whether some fundamental sharing of structural information between natural infection agents and artificial infection agents is at play here.

When the consensuses returned after the first alignment using ClustalW with Gonnet are given 3D *ab initio* realization, R1 (Figures 1 and 2) and R3 (Figures 5 and 6) produce models that most clearly distinguish between virus signature and worm signature consensuses, especially R3. The functional difference between a virus and a worm (the former requires human action, the latter self-propagates) can, for R3, be hypothesized to be biologically realized in the helical structure of virus signature consensus (Figure 5) in comparison to the sheeted structure of the worm signature consensus (Figure 6), for example.

When the rules for distinguishing between virus and worm signatures and extracted by PRISM from the doubly aligned sequences using R1 and R3 with ClustalW with Gonnet were given two forms of *ab initio* realization, almost identical meta-signature protein structures were produced (Figures 7 and 8), indicating that these meta-signatures share some fundamental properties despite their consensus differences. That is, the structures shown in Figures 7 and 8 provide a potential novel way to look at malware signatures irrespective of representation used. Some evidence for this comes from checking the meta-signature sequences against existing proteins, with matches found against GPCRs and Integrin A family signatures. GPCRs have been linked to cancer and metastasis (Dorsam and Gutkind, 2007). It is possible that further work will allow more explicit linkages to be made between uncontrolled growth of copies of mutant programs through malware infection and uncontrolled growth and spread of mutant copies of tumorous cells. Integrin also covers a family of proteins that adhere to cells and allow the transfer of information into and outside the cell. Adhesion to cells may be related to malware attaching itself to programs or email.

Finally, the experiments reported here contribute observations to the nascent area of malware theory. The fundamental theoretical question is whether there exists an algorithm that can take an arbitrary program and determine whether that program contains malware. Despite our experience and understanding of malware over the last 10 years, this is a surprisingly difficult theoretical question to answer for a number of reasons: getting hold of malware source code and uncertainty as to whether all the different types of malware logic have been identified are just two. More interestingly, malware often blurs the relationship between data and program. Parts of a program that look like data can be used to hide executable code. This complicates our understanding, from a theoretical computing perspective, of what counts as a 'program' in malware terms. Any new science or theory needs to be based on systematic observations of relevant phenomena so that theory development can be guided by evidence and data. The systematic experimentation described here has produced results using standard techniques that have a long history in bioinformatics and data mining, but these techniques can be effective if combined in novel ways to produce data relevant and useful for a possibly new area of computing and data mining called 'bio-inspired malware detection'. For instance, if more work is undertaken to build a library of 3D biologically-realized signatures of computer viruses and worms using the techniques described in this paper, it may be possible to construct an algorithm that identifies malware not just through text pattern matching of hexadecimal code but also through comparison conserved regions against known malware structures stored in databases in primary, secondary, tertiary and quaternary forms. Also, if it is shown that such malware structures share deep functional relationships with naturally occurring counterparts, biological knowledge could be adapted of how to stop, for example, adhesion or uncontrolled growth to inspire novel theoretical advances in our understanding of how to deal with computer malware detection and prevention.

**Acknowledgements**

The authors are grateful to Dr Ban Tao (National Institute of Information and Communications Technology, Koganei, Tokyo, Japan) and Dr Shaoning Pang (Department of Computing, Unitec Institute of Technology, Auckland, New Zealand) for their contributions to and inspiration for the initial parts of this research programme (please see Chen *et al*., 2012a and Chen *et al*., 2012b). The work reported in this paper is a significant departure from previously reported work and is the outcome of research conducted by the two contributing authors to this paper.

**Appendix A. Table 2: Overview of datasets produced by systems and methods**

| Method | Dataset | Description | Number |
|---|---|---|---|
|  | 60 virus signatures | Original hexadecimal signatures | 1 |
|  | 60 worm signatures | Original hexadecimal signatures | 1 |
| *Initial step* | VR1-VR3 | Original 60 virus hexadecimal signatures converted into amino acid representations 1-3 (Table 1) | 3 |
| *Initial step* | WR1-WR3 | Original 60 worm hexadecimal signatures converted into amino acid representations 1-3 (Table 1) | 3 |
| *Step (a)* | SAWCIR1-SAWCIR3 | Single aligned worm signatures using ClustalW, Identity matrix and Representations 1-3 | 3 |
| *Step (b)* | SAVCIR1-SAVCIR3 | Single aligned virus signatures using ClustalW, Identity matrix and Representations 1-3 | 3 |
| *Step (a)* | SAWCGR1-SAWCGR3 | Single aligned worm signatures using ClustalW, Gonnet matrix and Representations 1-3 | 3 |
| *Step (b)* | SAVCGR1-SAVCGR3 | Single aligned virus signatures using ClustalW, Gonnet matrix and Representations 1-3 | 3 |
| *Step (a)* | SAWCBR1-SAWCBR3 | Single aligned worm signatures using ClustalW, BLOSUM matrix and Representations 1-3 | 3 |
| *Step (b)* | SAVCBR1-SAVCBR3 | Single aligned virus signatures using ClustalW, BLOSUM matrix and Representations 1-3 | 3 |
| *Step (a)* | SAWTBR1-SAWTBR3 | Single aligned worm signatures using T-Coffee, BLOSUM matrix and Representations 1-3 | 3 |
| *Step (b)* | SAVTBR1-SAVTBR3 | Single aligned virus signatures using T-Coffee, BLOSUM matrix and Representations 1-3 | 3 |
| *Step (a)* | WCCIR1-WCCIR3 | Worm consensus using ClustalW, Identity matrix and Representations 1-3 | 3 |
| *Step (b)* | VCCIR1-VCCIR3 | Virus consensus using ClustalW, Identity matrix and Representations 1-3 | 3 |
| *Step (a)* | WCCGR1-WCGIR3 | Worm consensus using ClustalW, Gonnet matrix and Representations 1-3 | 3 |
| *Step (b)* | VCCGR1-VCCGR3 | Virus consensus using ClustalW, Gonnet matrix and Representations 1-3 | 3 |
| *Step (a)* | WCCBR1-WCCBR3 | Worm consensus using ClustalW, BLOSUM matrix and Representations 1-3 | 3 |
| *Step (b)* | VCCIR1-VCCIR3 | Virus consensus using ClustalW, BLOSUM matrix and Representations 1-3 | 3 |
| *Step (a)* | WCTBR1-WCTBR3 | Worm consensus using T-Coffee, BLOSUM matrix and Representations 1-3 | 3 |
| *Step (b)* | VCTBR1-VCTBR3 | Virus consensus using T-Coffee, BLOSUM matrix and Representations 1-3 | 3 |
| *Step (c)* | DACIR1-DACIR3 | Doubly aligned worm and virus signatures using Clustal, Identity matrix and Representations 1-3 (i.e. aligning SAWCIR1 with SAVCIR1, SAWCIR2 with SAVCIR2, SAWCIR3 with SAVCIR3) | 3 |
| *Step (c)* | DACGR1-DACGR3 | Doubly aligned worm and virus signatures using Clustal, Gonnet matrix and Representations 1-3 (i.e. aligning SAWCGR1 with SAVCGR1, etc., see above) | 3 |
| *Step (c)* | DACBR1-DACBR3 | Doubly aligned worm and virus signatures using Clustal, BLOSUM matrix and Representations 1-3 (i.e. aligning SAWCBR1 with SAVCBR1, etc., see above) | 3 |
| *Step (c)* | DATBR1-DATBR3 | Doubly aligned worm and virus signatures using T-Coffee, BLOSUM matrix and Representations 1-3 (i.e. aligning SAWTBR1 with SAVTBR1, etc., see above) | 3 |
| Total number of datasets |  |  | 68 |